\documentclass[pdflatex,sn-mathphys-num]{sn-jnl}

\usepackage{graphicx}%
\usepackage{multirow}%
\usepackage{amsmath,amssymb,amsfonts}%
\usepackage{amsthm}%
\usepackage{mathrsfs}%
\usepackage[title]{appendix}%
\usepackage{xcolor}%
\usepackage{textcomp}%
\usepackage{manyfoot}%
\usepackage{booktabs}%
\usepackage{algorithm}%
\usepackage{algorithmicx}%
\usepackage{algpseudocode}%
\usepackage{listings}%
\usepackage{tabularx}
\DeclareFontFamily{U}{rsfs}{\skewchar\font127 }
\DeclareFontShape{U}{rsfs}{m}{n}{%
<-6> rsfs5
<6-8> rsfs7
<8-> rsfs10
}{}
\newcolumntype{Y}{>{\centering\arraybackslash}X}

\theoremstyle{thmstyleone}%
\theoremstyle{thmstyletwo}%

\theoremstyle{thmstylethree}%

\begin{document}

\title[Energy Efficient Lung Tumor Segmentation with a modified CBAM ENet Skip Model]{Energy Efficient Lung Tumor Segmentation with a modified CBAM ENet Skip Model}


\author[1]{\fnm{Sharan} \sur{Thummagunti}}\email{tsharan76@gmail.com}
\equalcont{These authors contributed equally to this work.}

\author[1]{\fnm{Aayushi} \sur{Padia}}\email{aayushipadia1210@gmail.com}
\equalcont{These authors contributed equally to this work.}

\author[1]{\fnm{Abhiram} \sur{Koppuravuri}}\email{abhiramkoppuravuri@gmail.com}
\equalcont{These authors contributed equally to this work.}

\author[2]{\fnm{Prabhu} \sur{Prasad B M}}\email{prabhuprasad1990@gmail.com}

\author*[2]{\fnm{G N} \sur{Girish}}\email{girish.anit@gmail.com}

\affil[1]{\orgdiv{Department of Data Science and Artificial Intelligence}, \orgname{Indian Institute of Information Technology Dharwad}, \orgaddress{\street{Ittigatti Road}, \city{Dharwad}, \postcode{580009}, \state{Karnataka}, \country{India}}}

\affil[2]{\orgdiv{Department of Computer Science and Engineering}, \orgname{Indian Institute of Information Technology Dharwad}, \orgaddress{\street{Ittigatti Road}, \city{Dharwad}, \postcode{580009}, \state{Karnataka}, \country{India}}}


\abstract{Lung tumor segmentation plays a vital role in early detection and treatment planning of lung cancer. Although state-of-the-art deep learning models achieve high dice score coefficients (DSC), each forward pass requires computing large amounts of floating point operations, making them computationally expensive, requiring GPU acceleration, and having slow inference speed, limiting real-world clinical deployment. We propose a lightweight, energy-efficient model that leverages the ENet architecture for its real-time inference capability and low computational cost, enhanced with UNet-style skip connections between encoder and decoder stages for feature reuse and the lightweight Convolutional Block Attention Module (CBAM) for improved feature representation. The model is trained on publicly available data from The Cancer Imaging Archive (TCIA) via the Medical Segmentation Decathlon Challenge (MSD) and evaluated on the Lung Image Database Consortium and Image Database Resource Initiative (LIDC-IDRI) dataset. Our model achieves a DSC of 0.90812 (threshold of 0.5) with its compact architecture of 430,052 parameters and 0.549 Giga Multiply-Accumulate operations (GMAC), making it up to $10\times$faster, requires $79\times$ less GMAC than UNet, and has $22\times$ fewer parameters than SegNet. These characteristics make it well-suited for deployment in clinical settings where computational resources are limited and decrease the cost of utilizing such a system. Additionally, the explainable AI technique of Grad-CAM is used to provide visual interpretability of the model’s predictions and to identify regions where segmentation performance requires improvement.}




\maketitle
\section{Introduction}

Lung cancer or Bronchial Carcinoma remains one of the most fatal malignancies worldwide, owing to its aggressive nature and often diagnosed at advanced stages \cite{niranjana2017review}. Therefore, improving early detection is crucial, and Computed Tomography (CT), which is considered one of the best modalities for imaging, has emerged as a vital tool in this effort by offering detailed lung tissue imaging for analysis and identification of malignant tumors \cite{manikandan2019computer}. The advent of automated segmentation techniques that use deep learning models in Computer-Aided Diagnosis (CAD) systems has gained attention for reducing reliance on manual annotation and improving the consistency of tumor localization \cite{qureshi2023medical}. However, these models often rely on computationally expensive 3D convolution operations, which involve a high number of floating-point computations. As a result, they require GPU hardware acceleration for real-time inference on data \cite{xiao2020segmentation, zhang2022automatic, le2023rrc}, making them less feasible for deployment in environments with limited resources, such as small clinics or mobile health units, limiting their scalability and accessibility. Thus, energy-efficient neural networks which balance the inference speed, resource usage, and segmentation accuracy are critical for facilitating widespread adoption of AI-driven diagnostics in resource-constrained medical workflows.

In order to accurately segment the tumors, numerous neural network architectures for image segmentation have been proposed, such as SegNet \cite{chen2021segmentation, badrinarayanan2017segnet}, UNet and its variants \cite{saood2021covid, ronneberger2015u, cciccek20163d} or fully convolutional networks \cite{wang2017central, shen2017multi}. Moreover, the integration of transformer-based modules into encoder-decoder architectures such as UNETR \cite{hatamizadeh2022unetr} and Swin-UNETR \cite{hatamizadeh2021swin} has demonstrated improvements in Dice Score Coefficient (DSC) across various tasks. These powerful encoder-decoder architectures and the self-attention mechanism of transformers have large knowledge capacities and work with a high number of feature maps, enabling them to have strong feature representation capabilities. However, the base architecture of UNet and SegNet is based on the VGG-16 architecture \cite{simonyan2014very, muhtasim2024artificial}, a parameter-heavy model designed for multiclass image classification, leading to high Giga Multiply-Accumulate operations (GMAC) and long inference times. The integration of the transformer module, another computationally expensive and parameter-heavy architecture, adds to the existing high GMAC and inference time. In resource-constrained conditions, they become unsuitable for real-time use in resource-constrained settings.

Lightweight attention mechanisms have been developed as a promising way to enhance segmentation performance without additional computational overhead. The attention gate mechanism is a lightweight operation without adding any additional parameters and integrated in the UNet base architecture \cite{zhu2022attention} improving the representation power by reducing the focus on irrelevant target regions and enhancing spatial attention on relevant areas, but it does not account for inter-channel relationships. The Squeeze-and-Excitation module \cite{hu2018squeeze} introduced channel-wise attention using global average-pooling, but these features were argued to be suboptimal. The Convolutional Block Attention Module (CBAM) \cite{woo2018cbam} overcomes these by sequentially applying attention maps along both the channel and spatial dimensions and uses global max-pooling along with global average-pooling for channel attention. Understanding what is important in an image is achieved through the channel attention module while the spatial attention module utilizes inter-spatial relationships of features to focus on where the informative region lies. This design leads to an improved representation power of networks with minimal computational overhead. Due to the ease of seamlessly integrating CBAM into any Convolutional Neural Network (CNN) architecture with negligible overhead and can be trained jointly with the base CNN, it is an ideal choice for the proposed model. Furthermore, its original implementation with the ResNet architecture is structurally similar to the bottleneck modules in ENet, simplifying its integration into our proposed framework.

The ENet architecture was originally designed to overcome the limitation of traditional deep learning models that are computationally expensive and unable to perform real-time semantic segmentation on low-power mobile devices and autonomous vehicles \cite{paszke2016enet}. It is specifically optimized for low-latency and resource-constrained environments while maintaining competitive performance in terms of segmentation accuracy. Compared to UNet and SegNet, the ENet architecture offers a compelling balance between speed and accuracy. However, while it is computationally efficient, its segmentation accuracy can be further enhanced by improving its attention and feature re-usability capabilities. By integrating CBAM and UNet-style skip connections between encoder and decoder stages, the ENet architecture will have improved representation power and feature reuse through the skip connections. These skip connections also mitigate the need for pretraining the encoder to leverage transfer learning in order to fine-tune the decoder. Further discussion of this point is given in Section \ref{subsection:architecture}.

The main contribution of this research is an energy-efficient architecture for lung tumor segmentation by modifying the ENet architecture by adding UNet-style encoder-decoder skip connections and integrating CBAM into the encoder, improving feature reusability and representation without significant computational overhead, striking a strong balance between segmentation performance and computational efficiency. Demonstration of the model's effectiveness through comprehensive evaluation on the publicly available TCIA dataset \cite{prior2017public} through the Medical Segmentation Decathlon Challenge (MSD) and LIDC-IDRI dataset \cite{armato2011lung}, demonstrating that it significantly outperforms several state-of-the-art architectures without compromising inference speed and GMAC. The proposed architecture is up to 10\(\times\) faster, requires 79\(\times\) fewer GMAC than UNet and contains 22\(\times\) fewer parameters than SegNet, making it the ideal model for resource-constrained environments like mobile health units or rural clinical settings, reducing the cost of usage and eliminating the need for expensive GPU hardware acceleration.

\section{Related Work}

With ongoing advances in deep learning-based CAD systems having lead to the creation of numerous convolutional neural networks dedicated to lung tumor and nodule research, they predominantly explore three categories: 2D convolutional neural networks, 3D-based methods, and transformer-based methods.

\subsection{2D convolutional neural network}

The original U-Net architecture \cite{ronneberger2015u, rainio2025convolutional}, is a 2D encoder-decoder CNN with skip connections that enable high-resolution feature fusion for biomedical images. The encoder structure focuses on what features are present in the image, but due to downsampling, spatial information of where these features are present is lost. Hence, the decoder with feature reuse through skip connections from the encoder restores this spatial context, improving the segmentation accuracy of the model. Many subsequent 2D models inspired by this base architecture have been introduced with improvements. Attention UNet \cite{zhu2022attention} improves spatial attention in relevant areas by reducing the focus on irrelevant parts through its attention gate mechanism, UNet++ \cite{zhou2018unet++} introduces nested and dense skip connections across all levels between the encoder and decoder, providing more flexibility to utilize the features across various levels and thus narrowing the semantic gap between the encoder and decoder. However, a major drawback of 2D networks is the loss of contextual 3D spatial information, which is argued to lead to performance degradation. Since CT scans are inherently volumetric, slice-wise processing may ignore inter-slice continuity and introduce artifacts or inconsistencies in the segmented output. But working with the complete 3D volume is computationally expensive, with the 3D UNet itself having 542.19 GMAC for an input patch of 128 x 128 x 128. Moreover, even with a 2D architecture, the downsampling process exponentially increases the number of feature maps produced by the encoder after each convolution block, and the upsampling process doubles this number through the skip connections. Working with such a high number of feature maps leads to better performance with the trade-off being higher GMAC and more computation. On the other hand, the SegNet architecture mimics this downsampling process while opting to use max-unpooling instead of skip connections to restore lost spatial information \cite{badrinarayanan2017segnet}, allowing the model to work with a similar number of parameters and feature maps as UNet while requiring 1.6\(\times\) less GMAC than UNet.

\subsection{3D convolutional neural network}

To fully exploit the 3D information of the CT image, 3D UNet was introduced \cite{cciccek20163d, rainio2025convolutional}, extending the original UNet by replacing 2D convolutions and pooling operations with their 3D counterparts. This helps in learning the volumetric context, which is crucial to accurately delineating small nodules with subtle features in adjacent slices. Although 3D UNet offers performance improvements in volumetric segmentation, its increased memory footprint and training time pose practical limitations. In an effort to mitigate this, optimized 3D variants have emerged. 3D Res2UNet combines the benefits of Res2Net’s multi-scale feature learning and residual connectivity in a 3D setting \cite{xiao2020segmentation}. RRc-UNet 3D enhances 3D UNet with residual recalibration, enabling better feature reuse and representation learning \cite{le2023rrc}. nnU-Net \cite{ferrante2022application}, a self-configuring 3D pipeline, automatically adapts its network architecture and training to the target dataset, consistently ranking among the top performers in medical segmentation benchmarks. These models show better performance over their 2D counterparts, especially for small or complex lesions. For example, 3D-Res2UNet captures fine-grained structural information better than 2D variants owing to its deeper architecture and hierarchical feature extraction capability.

\subsection{Transformer architectures}

Recent development in deep learning for medical imaging has been significantly influenced by the success of Vision Transformers (ViTs), which have shown superior performance in capturing long-range dependencies in image data compared to traditional CNNs \cite{li2023tpfr}. Inspired by their success in natural image tasks, researchers have introduced transformer-based models to medical image segmentation, particularly for 3D volumetric data where global context is crucial. UNETR (U-Net with a ViT encoder for Medical Image Segmentation) \cite{hatamizadeh2022unetr} is a pioneering work that integrates pure transformers into the UNet framework, outperforming 3D UNet and nnU-Net in tasks such as spleen, liver, and brain tumor segmentation. Unlike the traditional UNet which relies solely on convolutional encoders, UNETR uses a ViT encoder that directly processes the volumetric input through non-overlapping 3D patches. The resulting patch embeddings are passed through multiple transformer layers, capturing global spatial dependencies across the entire volume. The decoder follows the UNet-style design and progressively upsamples the features to construct the final segmentation output. Skip connections between the transformer blocks and corresponding decoder blocks are present to fuse multi-scale semantic features. Swin-UNETR improves on UNETR with the Swin Transformer acting as the encoder \cite{hatamizadeh2021swin} and has faster convergence and reduced memory usage, due to efficient windowed attention. The Swin Transformer introduced the shifted window self-attention mechanism, which restricts computation of attention scores to non-overlapping windows while allowing interaction across windows through a shifting mechanism. This design brings linear computational complexity relative to image size, making it significantly more efficient for large 3D volumes and outperforming state-of-the-art models for brain tumor segmentation from MRI scans. However, transformer modules require high memory consumption during training and inference because of the computationally expensive transformer encoder and have slower inference speeds compared to CNN-based models.

\section{Methods}
\label{section:proposedwork}

\subsection{Dataset}

The Medical Segmentation Decathlon Challenge (MSD) is a collection of 10 different medical imaging segmentation challenges, each focusing on a specific organ or disease type. Task 6 of the Medical Segmentation Decathlon is dedicated to Lung Tumor Segmentation, which involves segmentation of lung tumors within CT scans. The following parameters were used to obtain the images: automatic tube current modulation range of 100–700 mA, a helical pitch, 0.9–1.0, a tube rotation speed of 0.5 s; section thickness below 1.5 mm; 120 kVp; and a sharp reconstruction kernel \cite{simpson2019large}. These volumes were acquired from The Cancer Imaging Archive (TCIA) \cite{prior2017public} and made publicly available through the MSD challenge.

\begin{figure}[!htbp]
\centering
\includegraphics[width=14 cm]{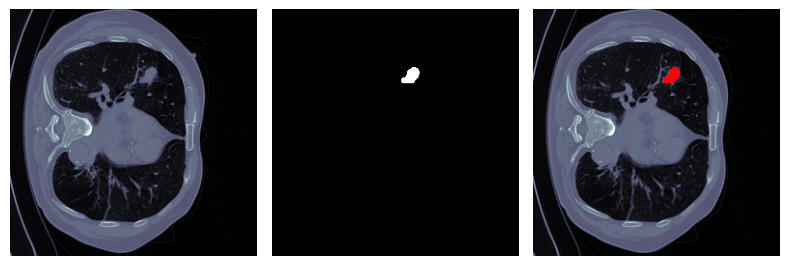}
\caption{2D slices from the MSD dataset with the CT slice (left), the ground truth segmentation mask (middle) and the overlap of both (right).\label{fig:output1}}
\end{figure}

To mitigate the computational expense of processing the whole 3D volume, each CT scan, with dimensions of 512 x 512 x X, where X is a variable that denotes the number of slices present in each CT scan, is converted into individual two-dimensional slices. These two-dimensional slices, as visualized in Figure \ref{fig:output1}, are used as the training and evaluation datasets. There are roughly 8.5 times more slices that do not contain a tumor, indicating a class imbalance of slices. Undersampling the majority class leads to a small dataset that does not represent the whole structure of the chest and lungs. Hence, oversampling of tumor slices is necessary to prevent the model from learning an all negative class prediction \cite{shorten2019survey}. This makes the model prone to overfitting, especially for attention-based architectures \cite{liu2022cerebrovascular, li2025ae} and requires data augmentation to bring more variety to existing data \cite{joffreysurvey}. Hence, it is necessary to select a wide variety of data augmentation techniques that complement each other and are suitable for the given type of disease, imaging modality and organ being diagnosed \cite{cossio2023augmenting, goceri2023medical, shorten2019survey}. A detailed explanation of the selected techniques is given in Section \ref{subsection:data}.

While the MSD dataset is used to train and validate the model performance, testing is performed on the entire Lung Image Database Consortium Image Collection (LIDC-IDRI) dataset. It consists of 1018 instances and 244,527 images stored in Digital Imaging and Communications in Medicine (DICOM), with a separate XML file associated with each volume that is produced after a two-phase image annotation process performed by four expert thoracic radiologists, necessary for creating the ground truth segmentation masks \cite{armato2011lung}. The test metrics obtained for each model are after comparing the model's predictions and the ground truth binary mask extracted from these XML files.

\begin{figure}[!htbp]
\centering
\includegraphics[width=14 cm]{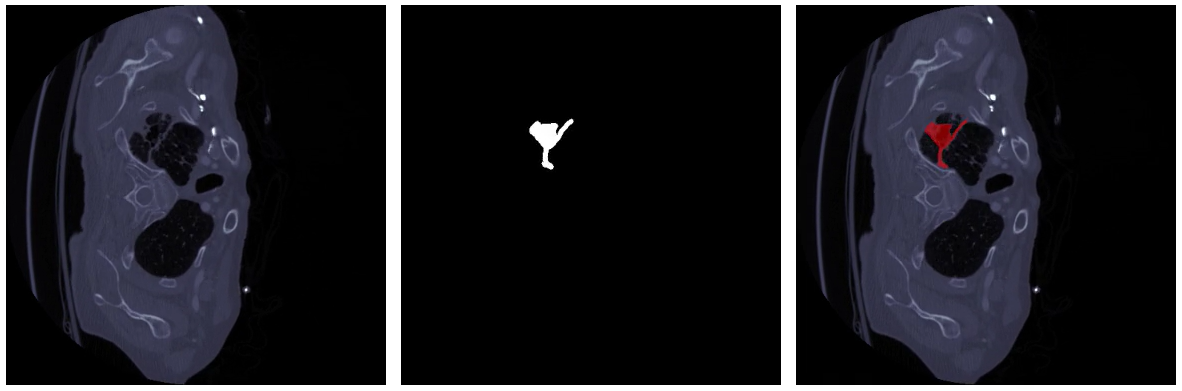}
\caption{2D slices from the LIDC-IDRI dataset with poor illumination conditions.\label{fig:output2}}
\end{figure}

Although the MSD and LIDC-IDRI dataset contain CT images, they exhibit notable differences in pixel intensity distribution and image quality. In the MSD dataset, the pixel intensities of a CT image are consistently bounded within a fixed range of \(-1024\) to \(3071\) Hounsfield Units (HU), while the LIDC-IDRI dataset contains a wider and more variable intensity range, with values extending from \(-3024\) up to \(3888\) HU. Additionally, MSD images are generally of higher definition and exhibit consistent lighting conditions, resulting in clearer and more visually coherent scans. In contrast, LIDC-IDRI images can vary significantly in both lighting and resolution, with some scans appearing pixelated or poorly lit with low contrast, making it difficult to analyze the image and accurately distinguish nodule boundaries. Figure \ref{fig:output2} visualizes a CT slice with suboptimal lighting, where even a large tumor is difficult to detect. To address these inconsistencies and ensure compatibility across datasets, all pixel intensities in CT images were clipped to the range of \(-1024\) to \(3071\) HU. Furthermore, data augmentation techniques that simulate variations in lighting and image quality were applied during training to improve model robustness and generalization across diverse imaging conditions. A detailed explanation of the pre-processing and data augmentation methods is provided in Section \ref{subsection:data}.

\subsection{Network Architecture}
\label{subsection:architecture}

\begin{figure}[!htbp]
\centering
\includegraphics[width=14 cm]{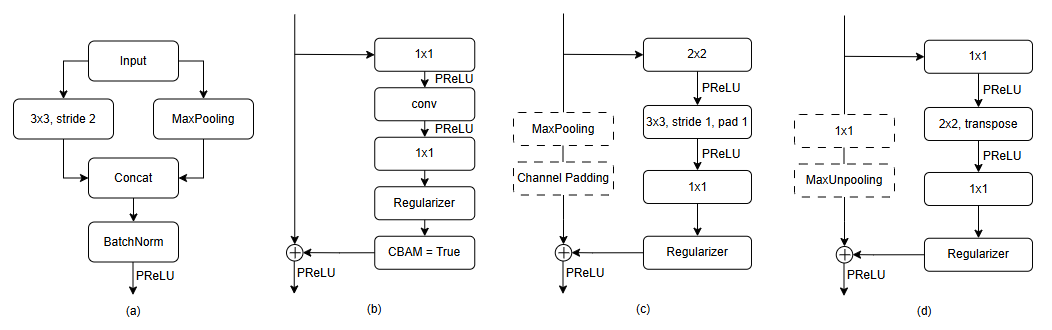}
\caption{(a) Initial block (b) Regular bottleneck module (c) Downsampling bottleneck module (d) Upsampling bottleneck module\label{fig:architecture}}
\end{figure}

The proposed network architecture is presented in Table \ref{tab:architecture} with expanded diagrams for each module in Figure \ref{fig:architecture}. The core design principle of the proposed modified ENet is early downsampling, followed by a deep but narrow network that significantly reduces the number of floating point operations for each forward pass, memory footprint, and inference time, without sacrificing accuracy. Its architecture is divided into several stages, as highlighted in the stage column.

\unskip

The initial block, as shown in Figure \ref{fig:architecture}a, processes the input images through a parallel max-pooling operation that results in 1 feature map and a convolution operation that results in 15 feature maps. The output of the convolution branch is concatenated with the output of the max-pooling branch responsible for preserving important spatial information. This final output passed through a batch normalization and activation layer. This architecture design is heavily inspired by the Inception architecture \cite{szegedy2016rethinking}.

The bottleneck modules are inspired by the ResNet architecture \cite{he2016deep}, with an extension branch consisting of convolutional filters and a main branch. These branches are then merged through element-wise addition, as shown in Figure \ref{fig:architecture}. This similarity simplifies the process of integrating CBAM as it is added after the convolution operations and before the residual connection in the form of the main branch is added. CBAM is integrated into all regular bottleneck modules of stage 2 and stage 3 of the architecture, improving the feature extraction capabilities on the encoder.

Each regular bottleneck module consists of three convolutional layers in the extension branch: a \(1\times1\) projection that decreases the feature maps by a factor of the internal ratio value, which is set to 4, a main convolutional layer (termed as 'conv' in Figure \ref{fig:architecture}b), and a \(1\times1\) expansion. The 'conv' layer is either a regular \(3\times3\) convolution, a dilated convolution \cite{ali2022efficient}, or a \(5\times5\) convolution decomposed into two \(5\times1\) and \(1\times5\) asymmetric convolutions. According to the integration flag passed to the regular bottleneck module, either CBAM or a skip connection is added.

In the downsampling block shown in Figure \ref{fig:architecture}c, an additional \(2\times2\) max-pooling layer with stride of 2 is added to the main branch followed by zero padding in the channel dimension. Zero padding ensures equal number of feature maps present in the main and extension branches in order to merge them through element-wise addition. The initial \(1\times1\) projection in the extension branch is replaced by a \(2\times2\) convolution with stride of 2. The 'conv' layer is replaced by a \(3\times3\) convolution with stride of 1 and 1 layer of padding. In the upsampling block shown in Figure \ref{fig:architecture}d, a \(1\times1\) convolution is performed, exchanging number of feature maps to increase the feature map size, followed by \(2\times2\) max-unpooling with stride of 2 in the main branch. The 'conv' layer from the extension branch is replaced with a \(2\times2\) transposed convolution of stride 2. Spatial dropout is used as the regularizer with \(p=0.01\) before bottleneck2.0 and \(p=0.1\) afterward. Batch normalization and activation layers are placed between all convolutions in a module. The encoder modules use PReLU as the activation, whereas the decoder modules use ReLU as the activation.

\begin{table}[!htbp] 
\caption{Model architecture for an input image resolution of \(1\times256\times256\).\label{tab:architecture}}
\begin{tabularx}{\textwidth}{ccYcc}
\toprule
\textbf{Name} & \textbf{Type} & \textbf{Integration} & \textbf{Output size} & \textbf{Stage} \\
\midrule
    initial & & & \(16\times128\times128\) & Initial \\
\midrule
    bottleneck1.0 & downsampling & & \(64\times64\times64\) & \multirow{5}{*}{Stage 1} \\
    bottleneck1.1 & & & \(64\times64\times64\) \\
    bottleneck1.2 & & & \(64\times64\times64\) \\
    bottleneck1.3 & & & \(64\times64\times64\) \\
    bottleneck1.4 & & & \(64\times64\times64\) \\
\midrule
    bottleneck2.0 & downsampling & & \(128\times32\times32\) & \multirow{9}{*}{Stage 2} \\
    bottleneck2.1 & & CBAM & \(128\times32\times32\) \\
    bottleneck2.2 & dilated & CBAM & \(128\times32\times32\) \\
    bottleneck2.3 & asymmetric & CBAM & \(128\times32\times32\) \\
    bottleneck2.4 & dilated & CBAM & \(128\times32\times32\) \\
    bottleneck2.5 & & CBAM & \(128\times32\times32\) \\
    bottleneck2.6 & dilated & CBAM & \(128\times32\times32\) \\
    bottleneck2.7 & asymmetric & CBAM & \(128\times32\times32\) \\
    bottleneck2.8 & dilated & CBAM & \(128\times32\times32\) \\
\midrule
    bottleneck3.0 & & CBAM & \(128\times32\times32\) & \multirow{8}{*}{Stage 3} \\
    bottleneck3.1 & dilated & CBAM & \(128\times32\times32\) \\
    bottleneck3.2 & asymmetric & CBAM & \(128\times32\times32\) \\
    bottleneck3.3 & dilated & CBAM & \(128\times32\times32\) \\
    bottleneck3.4 & & CBAM & \(128\times32\times32\) \\
    bottleneck3.5 & dilated & CBAM & \(128\times32\times32\) \\
    bottleneck3.6 & asymmetric & CBAM & \(128\times32\times32\) \\
    bottleneck3.7 & dilated & CBAM & \(128\times32\times32\) \\
\midrule
    bottleneck4.0 & upsampling & & \(64\times64\times64\) & \multirow{4}{*}{Stage 4} \\
    bottleneck4.1 & & Skip Connection (from bottleneck1.4) & \(64\times64\times64\) \\
    bottleneck4.2 & & & \(64\times64\times64\) \\
\midrule
    bottleneck5.0 & upsampling & & \(16\times128\times128\) & \multirow{2}{*}{Stage 5} \\
    bottleneck5.1 & & Skip Connection (from initial) & \(16\times128\times128\) \\
\midrule
    transpose conv & & & \(1\times256\times256\) \\
\bottomrule
\end{tabularx}
\end{table}

The initial stage contains a single initial block. Stage 1 consists of five bottleneck blocks, while stages 2 and 3 have the same structure, after excluding the first downsampling block of stage 2. The initial stage along with stages 1, 2 and 3 make up the network encoder, while stages 4 and 5 make up the network decoder. 

The indices of the max-pooling features are saved for the corresponding unpooling operations inspired by the SegNet architecture \cite{badrinarayanan2017segnet, rainio2025convolutional}, unlike the UNet architecture, which does not save the indices and relies on skip connections for spatial context. Rather than choosing one over the other, skip connections along with saving the indices for unpooling was chosen to provide maximum spatial information to the decoder, mitigating the need of pre-training the encoder to fine-tuning the decoder to focus on reconstructing this lost spatial context after performing max-unpooling. Without the use of skip connections, training the model end-to-end from random weight initializations leads to a plateau in performance for the first 100 epochs, after which a big jump in performance is observed. This is because the encoder has to first learn a strong representation of the data before the decoder can leverage this representation for spatial reconstruction and predicted segmentation. With the addition of skip connections, no such plateau and jump is observed. Moreover, the convergence time of the model is reduced by 32\%. The skip connection in bottleneck4.1 concatenates the output of stage 1 while in bottleneck5.1 it concatenates the output of the initial block. Integrating the attention gate mechanism \cite{zhu2022attention} into the encoder-decoder skip connections to enhance spatial attention does not improve the model's performance any further and only adds more parameters and computation to the model. Hence, CBAM provides enough channel and spatial attention, negating the need to integrate any additional attention mechanisms into the architecture.

\subsection{Model Training}

Training experiments are performed for 250 epochs on an input image size of \(1 \times 256 \times 256\) with batch size of 16. All finalized hyperparameters are listed in Table \ref{tab:hyperparameters}. A two-dimensional slice saved on the disk after undergoing data preparation is provided as input, from which the model provides the probability labels in the form of an output segmentation mask. All model outputs are stored in a list for every validation step, and at the end of every validation loop, these stored outputs are stacked, from which the evaluation metrics are calculated. More details on how the metrics are calculated is provided in Section \ref{subsection:metrics}.

\begin{table}[!htbp] 
\caption{Selected hyperparameter values for model training.\label{tab:hyperparameters}}
\begin{tabular}{@{}cc@{}}
\toprule
    \textbf{Hyperparameter} & \textbf{Value} \\
\midrule
    Image dimensions & \(256 \times 256\) \\
    Batch size & 16 \\ 
    No. of epochs & 250 \\
    Optimizer & Adam \\ 
    Learning rate & \(1 \times 10^{-4}\) \\ 
    Output activation head & Sigmoid \\ 
    Loss Function & Binary cross-entropy \\ 
\bottomrule
\end{tabular}
\end{table}

Binary cross-entropy, or log loss, is used because of its ability to provide flexibility in output probability predictions. The dice loss trains the model from an all positive prediction and to removing the pixels mislabeled as the positive class. This forces the model to output only extreme probability values, i.e. a high probability for tumors and a low probability for background. This makes the model perform better from a pure probability point of view, but the model will only assign a high probability to tumor pixels if it is strongly confident in its decision. Binary Cross Entropy, on the other hand, trains the model with flexibility as it allows the model to assign probability values from 0 to 1 for both the positive and background classes and only penalizes high probability misclassifications. This allows for probability output values according to the confidence level of the model. Due to such flexibility, the model could predict a tumor pixel with probability of 0.6 and when a threshold value is applied, classifying probabilities above the threshold as positive class and below as background class, provides an additional increase in the metric scores up to 6.5\%. This increase is not observed when the model is trained using dice loss because of the stricter prediction probabilities, leading to worse performance.

Unlike in the original ENet training process of pre-training the encoder before fine-tuning the decoder \cite{paszke2016enet}, transfer learning was not performed, and the model was trained from randomly initialized weights. Adam optimizer was selected with an initial learning rate of \(1\times10^{-4}\) through experimentation and research literature. A learning rate scheduler was used, which reduces the learning rate by a factor of 0.9 to \(1\times10^{-6}\) if the best validation DSC fails to improve in ten epochs. Moreover, the encoder-decoder skip connections in the architecture improve the gradient flow and feature reuse, further enhancing the performance.

\subsection{Data Preparation and Augmentation}
\label{subsection:data}

Before any preprocessing takes place, it is ensured that all CT scans followed the left, anterior, superior (L, A, S) alignment. After this assertion, the pixel intensities of each image are clipped to be in the range of 3071 to -1024 and standardized by dividing by 3071. Training the model using a lung window of 800 to -1024 degraded the segmentation performance. As the model should focus on lung tumors, the first 30 slices of the axial axis (which displays the images from the lower abdomen to the neck) consisting of parts of the lower abdomen are skipped for each volume, thereby reducing the complexity of the features to learn.

Due to the low number of slices that contain a tumor, it is important to choose the right data augmentation techniques that are known to work well in conjunction with CT scans of lung tumors. As the images are monochromatic, it is not possible to apply color space transformations like it would be with RGB images, which limits the options to geometric distortion, noise addition and contrast modification.

\begin{figure}[!htbp]
\centering
\includegraphics[width=10 cm]{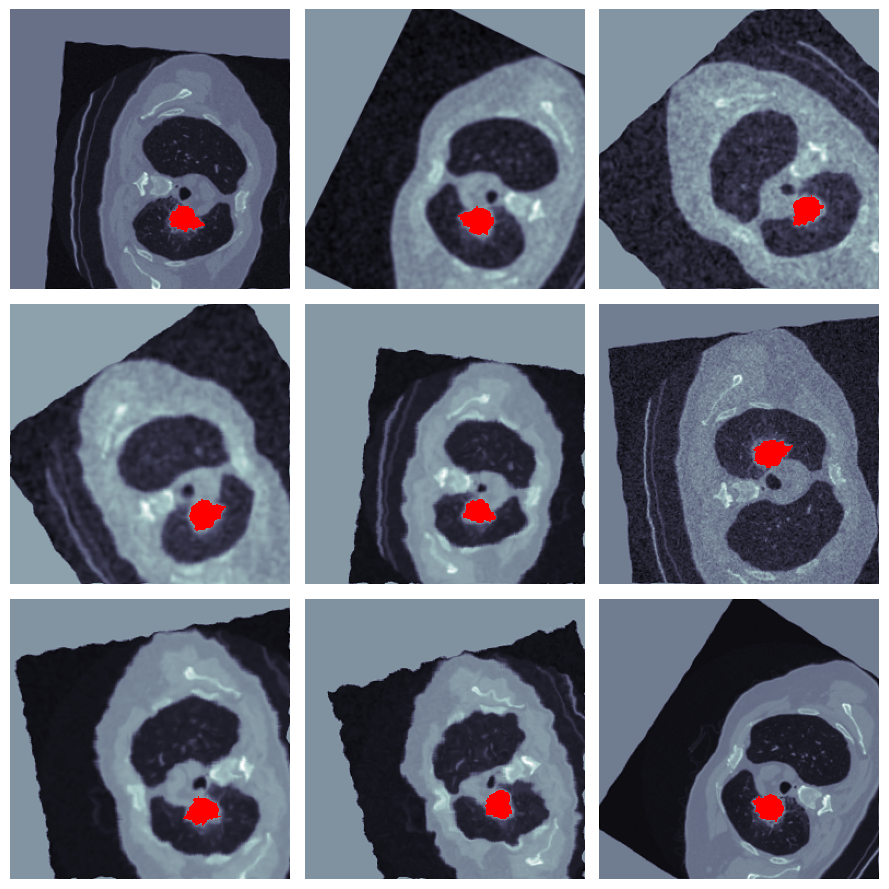}
\caption{A sample of 9 augmented images of a CT slice after passing through the data augmentation pipeline.\label{fig:output3}}
\end{figure}

Figure \ref{fig:output3} illustrates a sample of images after passing through the augmentation pipeline. The order of the transformations matters to avoid introducing unnecessary artifacts \cite{shorten2019survey}. Hence, the final pipeline for data augmentation is as follows: horizontal flipping, vertical flipping, linear contrast, additive gaussian noise, gaussian blur, affine transformations, and elastic deformation.

Experimentation results have shown that introducing Gaussian noise and Gaussian blur transformations into the pipeline helped reduce overfitting, as basic geometric transformations proved to not challenge the Spatial Attention Module (SAM) enough, leading to overfitting \cite{yu2024improved}. Vertical flipping is generally avoided as it can distort the anatomical orientation and introduce unrealistic scenarios, which can hinder model performance. However, because CT scans were reoriented to the (L, A, S) orientation, high angles of rotation and flipping would still result in a realistic valid image and introduce invariance to reflection \cite{hussain2018differential}, improving the model's robustness. Preserving the natural orientation would be mandatory when working with a 3D CNN but the advantage of working with 2D slices allows for such representation.

\section{Results}

\subsection{Evaluation Metrics}
\label{subsection:metrics}

The trained models are assessed using the evaluation metrics described below. \(\epsilon\) is a smoothing parameter with a value of \(10^{-8}\) used to avoid division by 0 when both predicted and ground truth masks contain no positive pixels, improving the numeric stability of the metrics.

\begin{itemize}

\item Dice Similarity Coefficient (DSC) - The DSC is a commonly used evaluation metric for semantic segmentation. It represents the extent of overlap between predicted segmentation mask and ground truth \cite{setio2016pulmonary, shorten2019survey}. The score value ranges from 0 to 1, where 0 represents no overlap and 1 represents perfect overlap. The formula for DSC is given in Equation \eqref{eqn:dice}:

\begin{equation}
    \text{Dice} = \frac{2 \cdot |Y \cap Y_{pred}| + \epsilon}{|Y| + |Y_{pred}| + \epsilon} = \frac{2 \cdot TP}{2 \cdot TP + FP + FN}
    \label{eqn:dice}
\end{equation}

\item Precision - It is defined as ratio of the sum of TPs to the total number of actual tumor pixels as given in Equation \eqref{eqn:precision}. It evaluates the model's ability to correctly identify tumor pixels in relation to all pixels it has classified as part of the tumor class.

\begin{equation}
    \text{Precision} = \frac{|Y \cap Y_{pred}| + \epsilon}{|Y_{pred}| + \epsilon} = \frac{TP}{TP + FP}
    \label{eqn:precision}
\end{equation}

\item Recall - It evaluates the model's ability to correctly detect all the pixels that belong to tumor class, relative to total number of pixels that actually belong to tumor class present in ground truth. The formula is given in Equation \eqref{eqn:recall}

\begin{equation}
    \text{Recall} = \frac{|Y \cap Y_{pred}| + \epsilon}{|Y| + \epsilon} = \frac{TP}{TP + FN}
    \label{eqn:recall}
\end{equation}

\item Intersection over Union (IoU) or Jaccard Index - It's used to measure accuracy of predicted segmentation with respect to the ground truth as given in Equation \eqref{eqn:iou}. It quantifies how well the predicted segmentation overlaps with the actual segmentation.

\begin{equation}
    \text{IoU} = \frac{|Y \cap Y_{pred}| + \epsilon}{|Y \cup Y_{pred}| + \epsilon} = \frac{TP}{TP + FP + FN}
    \label{eqn:iou}
\end{equation}

\end{itemize}

\begin{table}[!htbp]
  \centering
  \caption{Evaluation metric scores comparison over different architectures on the LIDC-IDRI dataset with threshold of 0.5.\label{tab:metricscores}}
  \begin{tabular}{@{}ccccc@{}}
    \toprule
    \textbf{Model} & \textbf{DSC} & \textbf{Precision} & \textbf{Recall} & \textbf{IoU} \\
    \midrule
    UNet \cite{ronneberger2015u} & 0.905826 & \textbf{0.926147} & 0.886377 & 0.827863 \\
    SegNet \cite{badrinarayanan2017segnet} & 0.900529 & 0.912512 & 0.888857 & 0.819057 \\
    ENet \cite{paszke2016enet} & 0.843147 & 0.880315 & 0.808991 & 0.728829 \\
    ENet + Skip & 0.845799 & 0.871146 & 0.821885 & 0.732800 \\
    Proposed model & \textbf{0.90812} & 0.917452 & \textbf{0.898976} & \textbf{0.831703} \\
    \bottomrule
  \end{tabular}
  \footnotetext{These model have been evaluated through experimentation.}
\end{table}

Table \ref{tab:metricscores} shows evaluation metrics scores of UNet, SegNet, baseline ENet, ENet with skip connections, and the proposed model tested on LIDC-IDRI. The proposed model outperforms state-of-the-art architectures, achieving a DSC of 0.90812 with a threshold of 0.5, demonstrating its ability to capture the entire nodule shape. Although UNet achieves the highest precision of 0.926147, the proposed model follows closely with a precision of 0.917452, indicating its strong performance in minimizing false positives. The proposed model also outperforms other architectures in recall (0.898976) and IoU (0.831703), further highlighting its robustness in segmentation. In comparison SegNet, ENet with skip connections and baseline ENet achieve DSC of 0.900529, 0.845799 and 0.843147. Testing on the LIDC-IDRI dataset confirms the capability and efficiency of the proposed architecture after thorough experimentation, and it can be extended to other medical segmentation tasks with appropriate task-specific modifications.

To evaluate whether the performance differences between the proposed model and UNet are statistically significant, a paired Wilcoxon signed-rank test \cite{woolson2007wilcoxon} is conducted on the DSC across the LIDC-IDRI volumes. The test resulted in $p = 0.02$ $(p > 0.05)$, indicating a statistically significant improvement. Although the absolute increase in DSC is modest, the result confirms that the gain is consistent across cases and is not attributable to random variation, reinforcing the practical value of our lightweight architecture.

\subsection{Model Performance}

This section presents the segmentation results on the LIDC-IDRI dataset and compares them with leading deep learning methods, as well as a discussion of its computational efficiency. All models have been trained using the pipeline discussed in Section \ref{section:proposedwork}.

\begin{figure}[!htbp]
\centering
\includegraphics[width=14 cm]{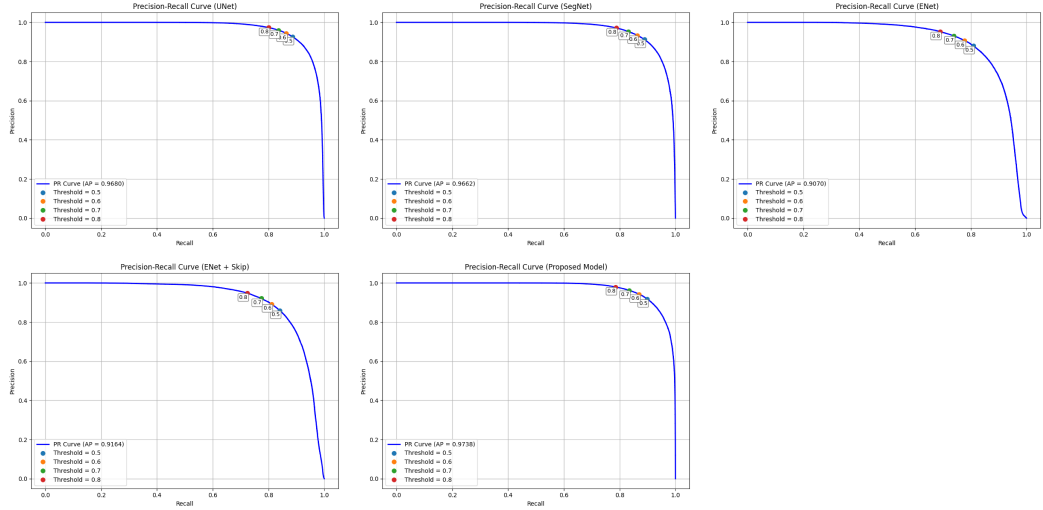}
\caption{Precision-Recall Curves with threshold overlay for different architectures. \label{fig:prcurve}}
\end{figure}   

\unskip

Figure \ref{fig:prcurve} shows the Precision-Recall (PR) curves and the corresponding Average Precision (AP) scores for the different model architectures, which are particularly informative for datasets with class imbalance. The models were ranked according to their AP scores, with the proposed model achieving the highest AP score of 0.9738, followed by UNet (0.9680), SegNet (0.9662), ENet with skip connections (0.9164) and baseline ENet (0.9070). The proposed model consistently maintains high precision and recall, even with higher threshold values such as 0.8, demonstrating its robustness in accurately identify tumors with minimal false positives. In contrast, both baseline ENet and ENet with skip connections experience a more gradual decline in precision at higher recall values, suggesting that while skip connections helped reduce training time, CBAM significantly enhanced the model's feature extraction capabilities.

\begin{table}[!htbp] 
\caption{Dice Score Coefficient (DSC) comparison with existing architectures. \label{tab:comparison}}
\begin{tabular}{@{}cc@{}}
\toprule
    \textbf{Model} & \textbf{DSC} \\ 
\midrule
    UNet with MobileNetV2 encoder \cite{riaz2023lung} & 0.8793 \\
    UNet with MobileNetV3 encoder \cite{suji2024exploring} \footnotemark[1] & 0.8824 \\ 
    UNet with EfficientNetB0 encoder \cite{suji2024exploring} \footnotemark[1] & 0.8503 \\ 
    nnU-Net \cite{ferrante2022application} & 0.9 \\ 
    3D-Res2UNet \cite{xiao2020segmentation} & \textbf{0.9530} \\ 
    RRc-Unet 3D \cite{le2023rrc} & 0.8777 \\ 
    UNETR \cite{hatamizadeh2022unetr} \footnotemark[1] & 0.5281 \\ 
    Swin-UNETR \cite{hatamizadeh2021swin} \footnotemark[1] & 0.6250 \\
    \textbf{Proposed model} \footnotemark[1] & \textbf{0.90812} \\
\bottomrule
\end{tabular}
\footnotetext[1]{These model have been evaluated through experimentation}
\end{table}

Table \ref{tab:comparison} compares the best DSC with the leading architectures. UNETR and Swin-UNETR struggle to achieve a high DSC as the vision transformer and swin transformer easily overfit on the training data, which has fewer tumor slices available within the dataset. Data augmentation was also not enough to prevent overfitting as the self-attention mechanism would negate the effects of any geometric transformations.

The proposed model beats the DSC of the leading architectures, with the exception of 3D-Res2UNet. This model was trained on a 70\% training split, a 20\% validation split, and a 10\% test split of the LUNA16 dataset \cite{armato2011lung}, benefiting from being trained on a larger subset of the LIDC-IDRI curated by slice-thickness criteria (\textgreater2.5 mm are excluded, yielding 888 scans). Taking into consideration that LUNA16 is a much larger dataset than the MSD dataset, the proposed model is still able to achieve a 0.90812 DSC. Moreover, The model maintains DSC \textgreater0.90 across datasets such as MSD and LIDC-IDRI without retraining, suggesting strong out-of-distribution generalization.

\begin{figure}[!htbp]
\centering
\includegraphics[width=14 cm]{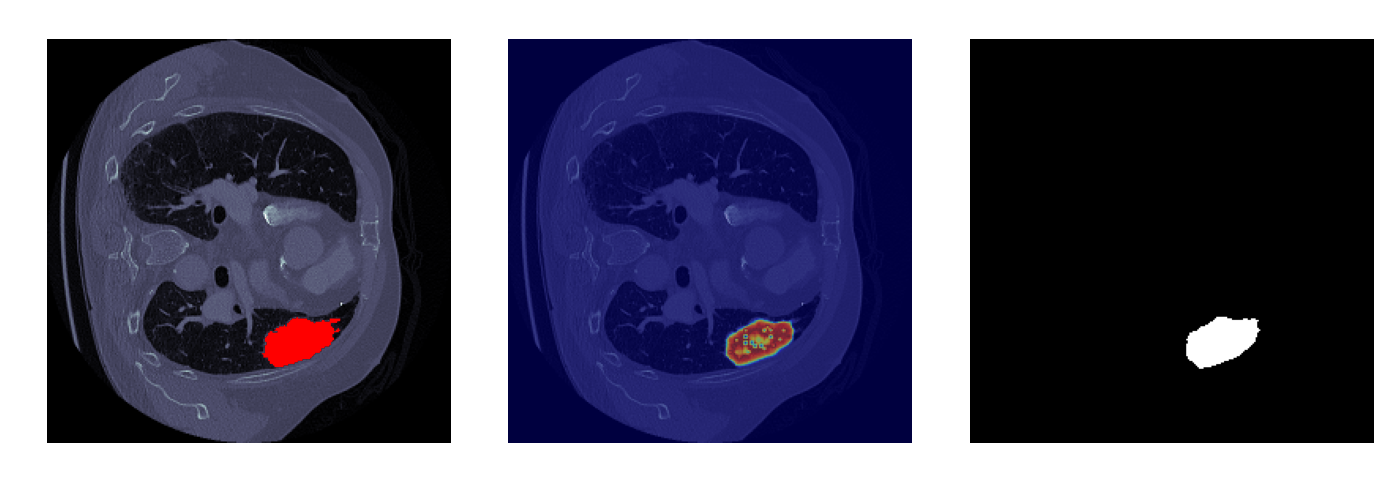}
\caption{Tumor slice with the ground truth mask overlapped (left), heatmap overlay of Grad-CAM (middle) and model segmentation mask prediction (right).\label{fig:gradcam}}
\end{figure}

To better understand the model, the explainable AI technique of Grad-CAM (Gradient-weighted Class Activation Mapping) \cite{xiao2021visualization} is used to enhance its interpretability and provide us with a visual understanding of its decision-making process. Grad-CAM enables us to visualize which areas of the CT scan captures the most attention due to the integration of CBAM. Figure \ref{fig:gradcam} gives us a deeper understanding of the output, where the model focuses more on the main area of the tumor but struggles to capture any thin outgrowths that branch away from the main tumor. This behavior is evident when a high threshold value, such as 0.8, is used, causing more false negatives in the model predictions and decreasing the recall. These observations highlight the importance of refining the model’s sensitivity to subtle tumor structures in future work either through architecture refinements or through model training.

\unskip

Finally, Table \ref{tab:computation} summarizes the parameter count and the computational cost of each architecture. Inference time is presented as the average time take for a model to process it's predictions for a single CT scan of dimension \(256\times256\times384\). The models were executed on an Intel Core i7-1165G7 CPU operating at a frequency of 2.8 GHz. 2D architectures have to process each slice sequentially, whereas 3D architectures have to process the CT scan by overlapping patch-based inference with a patch size of \(128\times128\times128\) and 50\% overlap between patches.

\begin{table}[!htbp] 
\caption{Number of parameters, computational efficiency and inference time on CPU to process a CT volume of dimensions \(256\times256\times384\).\label{tab:computation}}
\begin{tabular}{@{}lrrr@{}}
\toprule
    \textbf{Model} & \textbf{No. of parameters} & \textbf{GMAC} & \textbf{inference time (sec)} \\
\midrule    
    UNet \cite{ronneberger2015u} & 7,781,761 & 42.35 & 168.05 \\ 
    SegNet \cite{badrinarayanan2017segnet} & 9,374,593 & 26.68 & 93.94 \\
    UNet with MobileNetV2 encoder \cite{riaz2023lung} & 12,366,977 & 8.12 & 27.82 \\
    UNet with MobileNetV3 encoder \cite{suji2024exploring} & 9,370,561 & 7.17 & 22.59 \\ 
    UNet with EfficientNetB0 encoder \cite{suji2024exploring} & 14,233,597 & 8.12 & 31.29 \\
    3D-Res2UNet \cite{xiao2020segmentation} & 6,767,889 & 229.66 & 623.18 \\ 
    RRc-Unet 3D \cite{le2023rrc} & 22,576,259 & 800.88 & 1272.32 \\ 
    UNETR \cite{hatamizadeh2022unetr} & 115,654,577 & 4.58 & 133.73 \\ 
    Swin-UNETR \cite{hatamizadeh2021swin} & 6,302,203 & 3.69 & 109.42 \\
    ENet \cite{paszke2016enet} & 348,852 & 0.491 & 12.87 \\ 
    ENet + Skip & 360,628 & 0.547 & 14.34 \\ 
    \textbf{Proposed model} & 430,052 & 0.549 & 15.81 \\
\bottomrule
\end{tabular}
\end{table}

It is expected that 3D architectures have the highest GMAC and inference time due to how expensive 3D convolution operations are, leading to more multiply-accumulate operations in a single forward pass, with multiple forward passes due to patch-based inference, resulting in very high inference time to process a single CT scan. Among the 2D architectures, UNet has the highest GMAC and inference time. Although more computationally efficient variants of the UNet architecture, such as those adopting a pre-trained CNN encoder such as MobileNetV2, MobileNetV3 or EfficientNetB0, substantially reducing both GMAC and inference time, their efficiency gains remain far from the levels achieved by the proposed model. The ENet architecture is designed for computational efficiency, which is reflected in its extremely low GMAC and fast inference performance of 12.87 seconds. Integrating skip connections and CBAM increases GMAC and inference time, with the trade-off being improved segmentation accuracy yet even with these additions, the model remains significantly faster and lighter than all other architectures considered, removing the necessity for hardware acceleration in order to achieve real-time inference.

\subsection{Discussions}

Despite achieving a competitive Dice score value of 0.90812, beating architectures such as SegNet and UNet, recent advances such as transformer blocks, self-attention mechanisms, and the integration of other multimodal imaging in conjunction with CT scans have demonstrated significant improvements with DSC as high as 0.97 \cite{zhou2022apu}. Moreover, the Grad-CAM visualization of the model's decisions, as shown in Figure \ref{fig:gradcam}, reveals that the primary limitation of our model is that it struggles to capture any thin branches of tumor growth away from the main tumor volume with high levels of confidence. Training on a larger dataset, such as LUNA16, is expected to improve segmentation performance and lead to a higher DSC. Additionally, the use of 2D slices results in the loss of 3D context, which is argued to cause a loss of correlation between adjacent slices and reduce performance \cite{xiao2020segmentation}. Integrating 2.5D training by including adjacent slices above and below the target slice would include the benefits of 3D information and provide better edge detailing while retaining the benefits of low computational costs, potentially improving performance, especially for thin tumor branches \cite{kumar2024flexible}. Furthermore, the deliberate design choice of a compact architecture with strict GMAC constraints enables deployment on resource-constrained CAD systems without a hardware accelerator but imposes an upper bound on representational capacity compared to much larger transformer-based models.


\section{Conclusion}

In this paper, we present an energy-efficient deep learning model for lung tumor segmentation by modifying the ENet architecture through the integration of lightweight CBAM attention modules and UNet-style encoder-decoder skip connections. These enhancements strengthen feature representation and attention capabilities with minimal computational overhead. The results obtained by testing on the LIDC-IDRI dataset show that the proposed model achieves a DSC of 0.90812, precision of 0.917452, recall of 0.898976, and IoU of 0.831703 with a threshold of 0.5, outperforming architectures such as UNet and SegNet, while being up to 10\(\times\) faster, requiring 79\(\times\) fewer GMAC and having 22\(\times\) less parameters. The explainable AI technique of Grad-CAM visualization highlights the primary limitation as its inability to identify thin branches of tumor growth away from the main tumor volume with a high level of confidence.


%

\section*{Declarations}

\subsection*{Funding} The authors did not receive funding support from any organization for the submitted work.

\subsection*{Conflicts of Interests}

The authors declare that they have no conflict of interest.

\subsection*{Data Availability Statement}

Data sharing is not applicable to this article as no datasets were generated or analyzed during the current study. The dataset used in this work is available in the literature and we have used the dataset as cited by \cite{prior2017public}.

\subsection*{CRediT authorship contribution statement}

\textbf{Sharan Thummagunti, Ayushi Padia, and Abhiram Koppuravuri}: Methodology, Software, Validation, Investigation, Visualization, Writing - original draft.

\textbf{Prabhu Prasad B M}: Conceptualization, Validation, Investigation, Supervision, Methodology, Writing,
Review. 

\textbf{GN Girish}: Conceptualization, Methodology, Formal analysis, Project administration, Resources, Writing – review and editing, Supervision.

\bibliography{bibliography.bib}

@article{simpson2019large,
  title={A large annotated medical image dataset for the development and evaluation of segmentation algorithms},
  author={Simpson, Amber L and Antonelli, Michela and Bakas, Spyridon and Bilello, Michel and Farahani, Keyvan and Van Ginneken, Bram and Kopp-Schneider, Annette and Landman, Bennett A and Litjens, Geert and Menze, Bjoern and others},
  journal={arXiv preprint arXiv:1902.09063},
  year={2019}
}

@article{prior2017public,
  title={The public cancer radiology imaging collections of The Cancer Imaging Archive},
  author={Prior, Fred and Smith, Kirk and Sharma, Ashish and Kirby, Justin and Tarbox, Lawrence and Clark, Ken and Bennett, William and Nolan, Tracy and Freymann, John},
  journal={Scientific data},
  volume={4},
  number={1},
  pages={1--7},
  year={2017},
  publisher={Nature Publishing Group}
}

@article{armato2011lung,
  title={The lung image database consortium (LIDC) and image database resource initiative (IDRI): a completed reference database of lung nodules on CT scans},
  author={Armato III, Samuel G and McLennan, Geoffrey and Bidaut, Luc and McNitt-Gray, Michael F and Meyer, Charles R and Reeves, Anthony P and Zhao, Binsheng and Aberle, Denise R and Henschke, Claudia I and Hoffman, Eric A and others},
  journal={Medical physics},
  volume={38},
  number={2},
  pages={915--931},
  year={2011},
  publisher={Wiley Online Library}
}

@article{setio2016pulmonary,
  title={Pulmonary nodule detection in CT images: false positive reduction using multi-view convolutional networks},
  author={Setio, Arnaud Arindra Adiyoso and Ciompi, Francesco and Litjens, Geert and Gerke, Paul and Jacobs, Colin and Van Riel, Sarah J and Wille, Mathilde Marie Winkler and Naqibullah, Matiullah and S{\'a}nchez, Clara I and Van Ginneken, Bram},
  journal={IEEE transactions on medical imaging},
  volume={35},
  number={5},
  pages={1160--1169},
  year={2016},
  publisher={IEEE}
}

@article{shorten2019survey,
  title={A survey on image data augmentation for deep learning},
  author={Shorten, Connor and Khoshgoftaar, Taghi M},
  journal={Journal of big data},
  volume={6},
  number={1},
  pages={1--48},
  year={2019},
  publisher={Springer}
}

@article{cossio2023augmenting,
  title={Augmenting medical imaging: a comprehensive catalogue of 65 techniques for enhanced data analysis},
  author={Cossio, Manuel},
  journal={arXiv preprint arXiv:2303.01178},
  year={2023}
}

@article{goceri2023medical,
  title={Medical image data augmentation: techniques, comparisons and interpretations},
  author={Goceri, Evgin},
  journal={Artificial Intelligence Review},
  volume={56},
  number={11},
  pages={12561--12605},
  year={2023},
  publisher={Springer}
}

@article{liu2022cerebrovascular,
  title={Cerebrovascular segmentation model based on spatial attention-guided 3D inception U-Net with multi-directional MIPs},
  author={Liu, Yongwei and Kwak, Hyo-Sung and Oh, Il-Seok},
  journal={Applied Sciences},
  volume={12},
  number={5},
  pages={2288},
  year={2022},
  publisher={MDPI}
}

@misc{joffreysurvey,
  title={A survey on addressing high-class imbalance in big data. J Big Data (2018) 5: 42},
  author={Joffrey, L and Taghi, MK and Richard, B}
}

@inproceedings{hussain2018differential,
  title={Differential data augmentation techniques for medical imaging classification tasks},
  author={Hussain, Zeshan and Gimenez, Francisco and Yi, Darvin and Rubin, Daniel},
  booktitle={AMIA annual symposium proceedings},
  volume={2017},
  pages={979},
  year={2018}
}

@article{yu2024improved,
  title={An improved medical image segmentation framework with Channel-Height-Width-Spatial attention module},
  author={Yu, Xiang and Guo, Hongbo and Yuan, Ying and Guo, Wenjia and Yang, Xia and Xu, Hui and Kong, Yanqing and Zhang, Yudong and Zheng, Hairong and Li, Shengli},
  journal={Engineering Applications of Artificial Intelligence},
  volume={136},
  pages={108751},
  year={2024},
  publisher={Elsevier}
}

@article{paszke2016enet,
  title={Enet: A deep neural network architecture for real-time semantic segmentation},
  author={Paszke, Adam and Chaurasia, Abhishek and Kim, Sangpil and Culurciello, Eugenio},
  journal={arXiv preprint arXiv:1606.02147},
  year={2016}
}

@article{zhou2022apu,
  title={APU-Net: An Attention Mechanism Parallel U-Net for Lung Tumor Segmentation},
  author={Zhou, Tao and Dong, YaLi and Lu, HuiLing and Zheng, XiaoMin and Qiu, Shi and Hou, SenBao},
  journal={BioMed Research International},
  volume={2022},
  number={1},
  pages={5303651},
  year={2022},
  publisher={Wiley Online Library}
}

@inproceedings{xiao2021visualization,
  title={A visualization method based on the Grad-CAM for medical image segmentation model},
  author={Xiao, Mengying and Zhang, Liyuan and Shi, Weili and Liu, Jianhua and He, Wei and Jiang, Zhengang},
  booktitle={2021 International Conference on Electronic Information Engineering and Computer Science (EIECS)},
  pages={242--247},
  year={2021},
  organization={IEEE}
}

@article{manikandan2019computer,
  title={A computer-aided diagnosis system for lung cancer detection with automatic region growing, multistage feature selection and neural network classifier},
  author={Manikandan, T and Devi, B and Helanvidhya, T},
  journal={Int. J. Innov. Technol. Explor. Eng},
  volume={9},
  pages={409--413},
  year={2019}
}

@inproceedings{niranjana2017review,
  title={A review on image processing methods in detecting lung cancer using CT images},
  author={Niranjana, G and Ponnavaikko, M},
  booktitle={2017 international conference on technical advancements in computers and communications (ICTACC)},
  pages={18--25},
  year={2017},
  organization={IEEE}
}

@article{qureshi2023medical,
  title={Medical image segmentation using deep semantic-based methods: A review of techniques, applications and emerging trends},
  author={Qureshi, Imran and Yan, Junhua and Abbas, Qaisar and Shaheed, Kashif and Riaz, Awais Bin and Wahid, Abdul and Khan, Muhammad Waseem Jan and Szczuko, Piotr},
  journal={Information Fusion},
  volume={90},
  pages={316--352},
  year={2023},
  publisher={Elsevier}
}

@article{xiao2020segmentation,
  title={Segmentation of lung nodules using improved 3D-UNet neural network},
  author={Xiao, Zhitao and Liu, Bowen and Geng, Lei and Zhang, Fang and Liu, Yanbei},
  journal={Symmetry},
  volume={12},
  number={11},
  pages={1787},
  year={2020},
  publisher={MDPI}
}

@article{zhang2022automatic,
  title={Automatic lung tumor segmentation from CT images using improved 3D densely connected UNet},
  author={Zhang, Guobin and Yang, Zhiyong and Jiang, Shan},
  journal={Medical \& Biological Engineering \& Computing},
  volume={60},
  number={11},
  pages={3311--3323},
  year={2022},
  publisher={Springer}
}

@inproceedings{le2023rrc,
  title={RRc-UNet 3D for lung tumor segmentation from CT scans of Non-Small Cell Lung Cancer patients},
  author={Le, Van-Linh and Saut, Olivier},
  booktitle={Proceedings of the IEEE/CVF International Conference on Computer Vision},
  pages={2316--2325},
  year={2023}
}

@inproceedings{cciccek20163d,
  title={3D U-Net: learning dense volumetric segmentation from sparse annotation},
  author={{\c{C}}i{\c{c}}ek, {\"O}zg{\"u}n and Abdulkadir, Ahmed and Lienkamp, Soeren S and Brox, Thomas and Ronneberger, Olaf},
  booktitle={Medical Image Computing and Computer-Assisted Intervention--MICCAI 2016: 19th International Conference, Athens, Greece, October 17-21, 2016, Proceedings, Part II 19},
  pages={424--432},
  year={2016},
  organization={Springer}
}

@article{chen2021segmentation,
  title={Segmentation of lung computed tomography images based on SegNet in the diagnosis of lung cancer},
  author={Chen, Xiaodong and Duan, Qiongyu and Wu, Rong and Yang, Zehui},
  journal={journal of radiation research and applied sciences},
  volume={14},
  number={1},
  pages={396--403},
  year={2021},
  publisher={Elsevier}
}

@article{badrinarayanan2017segnet,
  title={Segnet: A deep convolutional encoder-decoder architecture for image segmentation},
  author={Badrinarayanan, Vijay and Kendall, Alex and Cipolla, Roberto},
  journal={IEEE transactions on pattern analysis and machine intelligence},
  volume={39},
  number={12},
  pages={2481--2495},
  year={2017},
  publisher={IEEE}
}

@article{saood2021covid,
  title={COVID-19 lung CT image segmentation using deep learning methods: U-Net versus SegNet},
  author={Saood, Adnan and Hatem, Iyad},
  journal={BMC Medical Imaging},
  volume={21},
  pages={1--10},
  year={2021},
  publisher={Springer}
}

@inproceedings{ronneberger2015u,
  title={U-net: Convolutional networks for biomedical image segmentation},
  author={Ronneberger, Olaf and Fischer, Philipp and Brox, Thomas},
  booktitle={Medical image computing and computer-assisted intervention--MICCAI 2015: 18th international conference, Munich, Germany, October 5-9, 2015, proceedings, part III 18},
  pages={234--241},
  year={2015},
  organization={Springer}
}

@article{wang2017central,
  title={Central focused convolutional neural networks: Developing a data-driven model for lung nodule segmentation},
  author={Wang, Shuo and Zhou, Mu and Liu, Zaiyi and Liu, Zhenyu and Gu, Dongsheng and Zang, Yali and Dong, Di and Gevaert, Olivier and Tian, Jie},
  journal={Medical image analysis},
  volume={40},
  pages={172--183},
  year={2017},
  publisher={Elsevier}
}

@article{shen2017multi,
  title={Multi-crop convolutional neural networks for lung nodule malignancy suspiciousness classification},
  author={Shen, Wei and Zhou, Mu and Yang, Feng and Yu, Dongdong and Dong, Di and Yang, Caiyun and Zang, Yali and Tian, Jie},
  journal={Pattern Recognition},
  volume={61},
  pages={663--673},
  year={2017},
  publisher={Elsevier}
}

@article{simonyan2014very,
  title={Very deep convolutional networks for large-scale image recognition},
  author={Simonyan, Karen and Zisserman, Andrew},
  journal={arXiv preprint arXiv:1409.1556},
  year={2014}
}

@inproceedings{hatamizadeh2022unetr,
  title={Unetr: Transformers for 3d medical image segmentation},
  author={Hatamizadeh, Ali and Tang, Yucheng and Nath, Vishwesh and Yang, Dong and Myronenko, Andriy and Landman, Bennett and Roth, Holger R and Xu, Daguang},
  booktitle={Proceedings of the IEEE/CVF winter conference on applications of computer vision},
  pages={574--584},
  year={2022}
}

@inproceedings{hatamizadeh2021swin,
  title={Swin unetr: Swin transformers for semantic segmentation of brain tumors in mri images},
  author={Hatamizadeh, Ali and Nath, Vishwesh and Tang, Yucheng and Yang, Dong and Roth, Holger R and Xu, Daguang},
  booktitle={International MICCAI brainlesion workshop},
  pages={272--284},
  year={2021},
  organization={Springer}
}

@inproceedings{szegedy2016rethinking,
  title={Rethinking the inception architecture for computer vision},
  author={Szegedy, Christian and Vanhoucke, Vincent and Ioffe, Sergey and Shlens, Jon and Wojna, Zbigniew},
  booktitle={Proceedings of the IEEE conference on computer vision and pattern recognition},
  pages={2818--2826},
  year={2016}
}

@inproceedings{woo2018cbam,
  title={Cbam: Convolutional block attention module},
  author={Woo, Sanghyun and Park, Jongchan and Lee, Joon-Young and Kweon, In So},
  booktitle={Proceedings of the European conference on computer vision (ECCV)},
  pages={3--19},
  year={2018}
}

@article{zhu2022attention,
  title={Attention-Unet: A deep learning approach for fast and accurate segmentation in medical imaging},
  author={Zhu, Ziyi and Yan, Yafeng and Xu, Ruilin and Zi, Yun and Wang, Jinyin},
  journal={Journal of Computer Science and Software Applications},
  volume={2},
  number={4},
  pages={24--31},
  year={2022}
}

@inproceedings{hu2018squeeze,
  title={Squeeze-and-excitation networks},
  author={Hu, Jie and Shen, Li and Sun, Gang},
  booktitle={Proceedings of the IEEE conference on computer vision and pattern recognition},
  pages={7132--7141},
  year={2018}
}

@inproceedings{zhou2018unet++,
  title={Unet++: A nested u-net architecture for medical image segmentation},
  author={Zhou, Zongwei and Rahman Siddiquee, Md Mahfuzur and Tajbakhsh, Nima and Liang, Jianming},
  booktitle={Deep learning in medical image analysis and multimodal learning for clinical decision support: 4th international workshop, DLMIA 2018, and 8th international workshop, ML-CDS 2018, held in conjunction with MICCAI 2018, Granada, Spain, September 20, 2018, proceedings 4},
  pages={3--11},
  year={2018},
  organization={Springer}
}

@article{ferrante2022application,
  title={Application of nnU-Net for automatic segmentation of lung lesions on CT images and its implication for radiomic models},
  author={Ferrante, Matteo and Rinaldi, Lisa and Botta, Francesca and Hu, Xiaobin and Dolp, Andreas and Minotti, Marta and De Piano, Francesca and Funicelli, Gianluigi and Volpe, Stefania and Bellerba, Federica and others},
  journal={Journal of clinical medicine},
  volume={11},
  number={24},
  pages={7334},
  year={2022},
  publisher={MDPI}
}

@inproceedings{he2016deep,
  title={Deep residual learning for image recognition},
  author={He, Kaiming and Zhang, Xiangyu and Ren, Shaoqing and Sun, Jian},
  booktitle={Proceedings of the IEEE conference on computer vision and pattern recognition},
  pages={770--778},
  year={2016}
}

@article{riaz2023lung,
  title={Lung tumor image segmentation from computer tomography images using MobileNetV2 and transfer learning},
  author={Riaz, Zainab and Khan, Bangul and Abdullah, Saad and Khan, Samiullah and Islam, Md Shohidul},
  journal={Bioengineering},
  volume={10},
  number={8},
  pages={981},
  year={2023},
  publisher={MDPI}
}

@article{suji2024exploring,
  title={Exploring pretrained encoders for lung nodule segmentation task using lidc-idri dataset},
  author={Suji, R Jenkin and Godfrey, W Wilfred and Dhar, Joydip},
  journal={Multimedia Tools and Applications},
  volume={83},
  number={4},
  pages={9685--9708},
  year={2024},
  publisher={Springer}
}

@article{woolson2007wilcoxon,
  title={Wilcoxon signed-rank test},
  author={Woolson, Robert F},
  journal={Wiley encyclopedia of clinical trials},
  pages={1--3},
  year={2007},
  publisher={Wiley Online Library}
}

@article{kumar2024flexible,
  title={A flexible 2.5 D medical image segmentation approach with in-slice and cross-slice attention},
  author={Kumar, Amarjeet and Jiang, Hongxu and Imran, Muhammad and Valdes, Cyndi and Leon, Gabriela and Kang, Dahyun and Nataraj, Parvathi and Zhou, Yuyin and Weiss, Michael D and Shao, Wei},
  journal={Computers in Biology and Medicine},
  volume={182},
  pages={109173},
  year={2024},
  publisher={Elsevier}
}

@article{rainio2025convolutional,
  title={Convolutional neural networks for tumor segmentation by cancer type and imaging modality: a systematic review},
  author={Rainio, Oona and Kl{\'e}n, Riku},
  journal={Network Modeling Analysis in Health Informatics and Bioinformatics},
  volume={14},
  number={1},
  pages={58},
  year={2025},
  publisher={Springer}
}

@article{ali2022efficient,
  title={An efficient U-Net framework for lung nodule detection using densely connected dilated convolutions},
  author={Ali, Zeeshan and Irtaza, Aun and Maqsood, Muazzam},
  journal={The Journal of Supercomputing},
  volume={78},
  number={2},
  pages={1602--1623},
  year={2022},
  publisher={Springer}
}

@article{li2023tpfr,
  title={TPFR-Net: U-shaped model for lung nodule segmentation based on transformer pooling and dual-attention feature reorganization},
  author={Li, Xiaotian and Jiang, Ailian and Qiu, Yanfang and Li, Mengyang and Zhang, Xinyue and Yan, Shuotian},
  journal={Medical \& Biological Engineering \& Computing},
  volume={61},
  number={8},
  pages={1929--1946},
  year={2023},
  publisher={Springer}
}

@article{muhtasim2024artificial,
  title={Artificial intelligence for detection of lung cancer using transfer learning and morphological features},
  author={Muhtasim, Nafe and Hany, Umma and Islam, Tahmina and Nawreen, Nusrat and Mamun, Abdullah Al},
  journal={The Journal of Supercomputing},
  volume={80},
  number={10},
  pages={13576--13606},
  year={2024},
  publisher={Springer}
}

@article{li2025ae,
  title={AE-UNet: a composite lung CT image segmentation framework using attention mechanism and edge detection},
  author={Li, Hongzhi and Ren, Zhanghao and Zhu, Guoqing and Wang, Jiaxi},
  journal={The Journal of Supercomputing},
  volume={81},
  number={1},
  pages={331},
  year={2025},
  publisher={Springer}
}

\end{document}